\documentclass[letterpaper, 10 pt, conference]{ieeeconf}  

\IEEEoverridecommandlockouts                              

\usepackage{cite}
\let\labelindent\relax
\usepackage{enumitem}
\usepackage{amsmath,amssymb,amsfonts}
\usepackage{graphicx}
\usepackage{textcomp}
\usepackage{comment}
\usepackage{subfig}
\usepackage{siunitx}
\usepackage{xcolor}
\usepackage{caption}
\usepackage{array}
\usepackage{algorithm}
\makeatletter
\usepackage{algpseudocode}
\usepackage{soul}
\usepackage[capitalise]{cleveref}
\DeclareCaptionFont{ninept}{\fontsize{9pt}{11pt}\selectfont #1}
 
\title{\LARGE \bf
Learning an Interpretable Model for Driver Behavior Prediction \\ with Inductive Biases}

\author{Salar Arbabi, Davide Tavernini, Saber Fallah and Richard Bowden
	\thanks{Salar Arbabi, Davide Tavernini and Saber Fallah are with the Centre for Automotive Engineering, University of Surrey, Guildford, GU2 7XH, U.K. (e-mail: \{s.arbabi, d.tavernini, s.fallah\}@surrey.ac.uk).}
	\thanks{Richard Bowden is with the Centre for Vision, Speech and Signal Processing, University of Surrey, Guildford GU2 7XH, U.K. (e-mail: r.bowden@surrey.ac.uk).}
}

\begin{document}

\maketitle
\thispagestyle{empty}
\pagestyle{empty}

\begin{abstract}
To plan safe maneuvers and act with foresight, autonomous vehicles must be capable of accurately predicting the uncertain future.  
In the context of autonomous driving, deep neural networks have been successfully applied to learning predictive models of human driving behavior from data. However, the predictions suffer from cascading errors, resulting in large inaccuracies over long time horizons. 
Furthermore, the learned models are black boxes, and thus it is often unclear how they arrive at their predictions.  
In contrast, rule-based models---which are informed by human experts---maintain long-term coherence in their predictions and are human-interpretable.
However, such models often lack the sufficient expressiveness needed to capture complex real-world dynamics. 
In this work, we begin to close this gap by embedding the Intelligent Driver Model, a popular hand-crafted driver model, into deep neural networks. 
Our model's transparency can offer considerable advantages, e.g., in debugging the model and more easily interpreting its predictions. 
We evaluate our approach on a simulated merging scenario, showing that it yields a robust model that is end-to-end trainable and provides greater transparency at no cost to the model's predictive accuracy.
\end{abstract}

\section{INTRODUCTION}
 
In a future society where autonomy becomes ubiquitous, one can imagine robots operating in complex, human-dominated environments, such as self-driving cars driving amidst human-driven vehicles on the road.  
Safe and reliable integration of robots into daily life demands that they account for the potential impact of their actions on the world. As an example, consider merging into a high-density lane of slow-moving traffic. In the same scenario, an autonomous car may deem all paths forward as unsafe and become unable to progress if it cannot anticipate that due to its actions, other drivers might cooperate and yield.
To this end, a robot should have a mental model of the world through which it can explore a set of possible decision options in its imagination, and learn how to act from the imagined~outcomes \cite{hafner2020mastering}. 
 
We posit that a predictive model should exhibit several characteristics, which in turn will drive the choice of model architecture and model learning strategy. First, in highly dynamic environments, actions may have long-term consequences that must be considered by the robot; being reactive to immediate situations may yield a ``myopic'' robot that lacks foresight. Thus, the model predictions should maintain long-term coherence, such that the robot can imagine consequence of its actions on distant futures. Second, given that future is often riddled with uncertainty, the model should capture the uncertainty about the world. 
Finally, given the safety-critical nature of many tasks involving human-robot interactions, interpretable representations that can explain the model and its predictions and help to identify failure modes are crucial.

\begin{figure}
    \centering
    \includegraphics[trim={0cm 15.5cm 21cm 0cm}, clip=true,
    width=\linewidth]{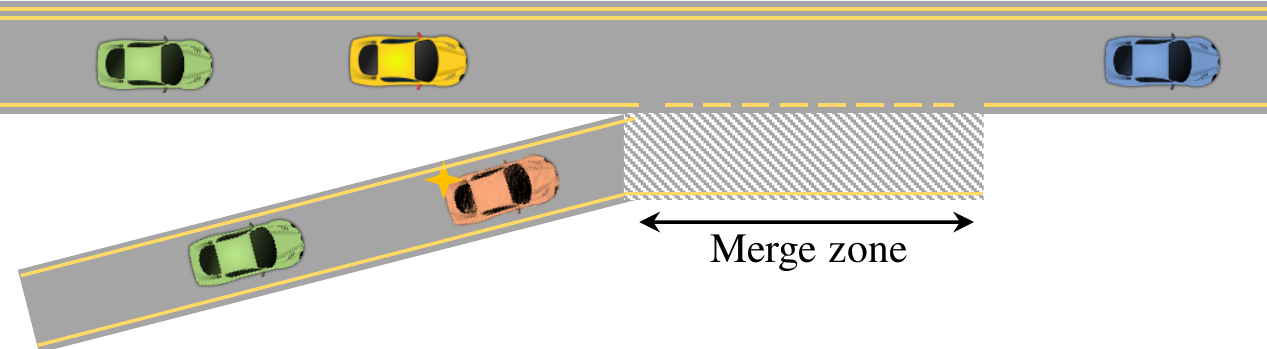}
    \caption{Example of a merging scenario. The merging vehicle (orange) is an autonomous car. The vehicles on the main road can yield to the autonomous car, or they can ignore it. We develop a driver model to predict the future behavior of the vehicles on the main road.}\label{fig:scene_schematic}
\end{figure}

In this paper, we use autonomous driving as our use case, where the mental model constitutes driving agents whose behavior we aim to predict. 
We develop our approach around a merging scenario (see \cref{fig:scene_schematic}), where we aim to predict the behavior of the drivers driving on the main road. The main-road drivers can choose whether to attend and yield to the approaching vehicles on the on-ramp and as a result,
it is necessary for us to estimate their driving intent. 

Deep neural networks (DNNs) have shown great potential in modeling complex dynamics and enabling the learning of driver models from naturalistic driving data. 
Although the successful application of DNNs to driver behavior modeling has been demonstrated \cite{casas2020implicit, casas2019spatially, casas2020importance, bhattacharyya2019simulating, henaff2019model, suo2021trafficsim, morton2017simultaneous}, substantial challenges remain. One major problem is \textit{distribution shift}: the tendency of such models to produce 
increasingly suboptimal actions, at times expressing dangerous behaviors (e.g., 
driving off the road, colliding with others, or braking too hard \cite{bhattacharyya2019simulating}) when exposed to scenarios with insufficient training data.
This problem is exacerbated in model-based planning, as an agent is likely to imagine itself in novel states when exploring different decision options\cite{kober2013reinforcement}.  

In addition to the problem of distribution shift, the strength of DNNs comes at the cost of the low interpretability of their black-box representations. The learned models can have many thousands of parameters, offering little to no explanation/visibility into how they arrive at their predictions.  
While research on model interpretability has been growing rapidly in recent years \cite{gilpin2018explaining}, most methods are confined to domains with data types that are easily interpretable to humans: image and natural language processing. 
In contrast, driver behavior models derive their predictions from noisy numeric observations of the environment, which are hardly human-interpretable.

In this paper, we embed a task-specific structure in deep networks that is amenable to gradient-based optimization. The imposed structure captures domain knowledge and manifests as a semantic layer at the output of the network, mapping arbitrary outputs of previous layers to driver actions.  
One domain knowledge with regards to driving is that human drivers tend to avoid collisions, something difficult to reason about for classic imitation learning, as such rare cases are under-represented in training data. 
In addition to structure embedding, we leverage the framework of conditional variational autoencoders (CVAE) \cite{sohn2015learning} to infer the unobserved (latent) states (e.g., intentions and dispositions) of drivers from their past motion; this enables our model to output driver actions that are conditioned on sampled latent states, capturing the variation in drivers' behavior and providing uncertainty estimates for the predictions.  

\section{RELATED WORKS}

\subsection{Inductive Biases in Deep Learning}
Recent works have shown that imposing well-motivated inductive biases on neural network architectures can help with learning more interpretable representations \cite{cranmer2020discovering} and facilitate generalization \cite{wilson2020bayesian}. 
In addition to the network architecture, inductive biases can also be incorporated through the learning objective and structure embeddings. 
Some works related to learning driver models include expert-informed terms in the objective, such as loss penalties for causing collisions \cite{casas2020importance, bhattacharyya2019simulating, henaff2019model, suo2021trafficsim}. There have also been several attempts at directly imposing the structure of a dynamical system on DNNs for robot control\cite{bahl2020neural, gupta2020structured}. Unlike the dominant paradigm of learning policies in raw action spaces, they propose the re-parameterization of the robot's system dynamics via deep neural network-based policies, achieving significant gains in policy performance. 

\subsection{Interpretable Driver Models}
One commonly used driver model is the Intelligent Driver Model (IDM) \cite{treiber2000congested}, which predicts the longitudinal acceleration of a vehicle based on the traffic state. The design of IDM is motivated by the knowledge that a typical driver's goal is to drive safely and efficiently. Rule-based models such as IDM generate plausible traffic flow and have a few interpretable parameters that determine the model's output.
Offline calibration of IDM parameters has been performed in existing work \cite{punzo2014we}. 
However, these methods yield ``average'' parameters for a population of drivers, and as such cannot capture the idiosyncrasies of individual drivers present in real-world driving. Existing work has also performed online parameter estimation for the IDM using an Extended Kalman filter \cite{monteil2015real}, nonlinear least squares \cite{ward2017probabilistic} and particle filtering \cite{sunberg2020improving, bhattacharyya2020online}. Put simply, they treat the problem of behavior prediction as an online learning problem, stating: \textit{what are the IDM parameters that best describe the observed vehicle trajectory?}
The particle filtering approach proposed by Bhattacharyya et al. \cite{bhattacharyya2020online} was evaluated on a real-world driving dataset, achieving superior performance compared to several deep-learning-based models.  
While the approach takes advantage of the IDM's model structure, it requires explicit knowledge of the observation and motion models; this restricts the application of their approach to car-following scenarios, where an IDM vehicle is assumed to always pay attention to its front neighbor.

\section{Model Formulation}

\begin{figure*}
    \centering
    \includegraphics[trim={0cm 10.3cm 4cm 0cm}, clip=true,
    width=\linewidth]{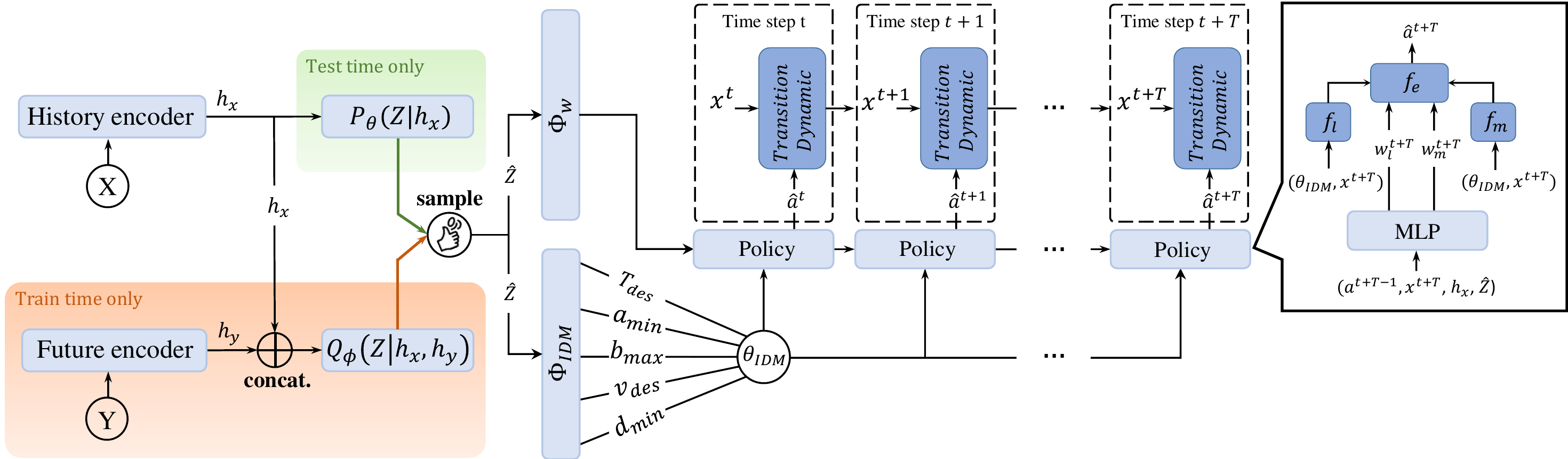}
    \caption{Overview of the architecture for NIDM. During model training, we draw samples $\hat Z \sim  Q_{\phi}(Z|h_x, h_y)$. At test time, we draw samples $\hat Z \sim P_{\theta}(Z|h_x)$ instead. The part of the model denoted by $\Phi_{\mathrm{IDM}}$ reasons in the lower dimensional space of parameters $\theta_{\mathrm{IDM}}$ which are then used within the unrolled policy to predict driver actions.}\label{fig:model_architecture}
\end{figure*}

We follow the structure imposed by IDM to determine a driver's longitudinal acceleration. Given a set of driver parameters and the traffic state $x^t$, IDM outputs the driver accelerations over time that provide a trade-off between a driver's goal of reaching a desired speed and keeping a safe distance from the vehicle ahead. 
The set of driver parameters $\theta_{\mathrm{IDM}} = \{v_{\mathrm{des}}, d_{\mathrm{min}}, T_{\mathrm{des}}, a_{\mathrm{max}}, b_{\mathrm{max}}\}$
represent the driver's disposition which consists of the driver's desired speed $v_{\mathrm{des}}$, the minimum separation distance $d_{\mathrm{min}}$, the time gap $T_{\mathrm{des}}$, the ideal maximum acceleration $a_{\mathrm{max}}$ and the ideal maximum deceleration $b_{\mathrm{max}}$. 

A driver's longitudinal acceleration is determined by
\begin{equation}  \label{equ:idm_action}
\begin{aligned} 
f_{\mathrm{IDM}} = a_{\mathrm{max}}\left(1-\left(\frac{v^t}{v_{\mathrm{des}}}\right)^4-\left(\frac{d_{\mathrm{des}}(v^t, \Delta v^t)}{d^t}\right)^2\right)
\end{aligned}
\end{equation}
where $v^t$ is the vehicle speed, and $\Delta v^t$ and $d^t$ are the relative speed (approach rate) and distance headway to the vehicle ahead, respectively. The desired distance $d_{\mathrm{des}}$ is given by
\begin{equation} \label{equ:des_d}
\begin{aligned} 
d_{\mathrm{des}}(v^t, \Delta v^t) = d_{\mathrm{min}} + T_{\mathrm{des}}v^t + \frac{v^t\Delta v^t}{2\sqrt{a_{\mathrm{max}}b_{\mathrm{max}}}}
\end{aligned}
\end{equation}

We additionally introduce the attention parameters $w_l$ and $w_m$ to model the degree of a driver's attentiveness towards its front neighbor and a vehicle on the on-ramp, respectively. 
For a driver who is attending to the vehicle on the on-ramp and is being cooperative, $w_m\approx1$, while $w_m$ approaches zero for a non-cooperative driver. 
We will refer to the parameters $\{\theta_{\mathrm{IDM}}, w_l, w_m\}$ as the driver's internal state, emphasizing that they are not directly observable and can only be estimated. 
We note that our formulation can be extended to accommodate scenarios with more interacting vehicles by considering additional attention parameters.
We next describe how to combine the IDM's dynamic structure within a deep network as a way of incorporating inductive biases in the learned model. 

\section{Neural Intelligent Driver Model}

In this section, we introduce a policy network called the Neural Intelligent Driver Model (NIDM) that, given the vehicles' motion history, produces distributions over their sequence of future actions. 

\subsection{Neural Network Parameterization of NIDM}

We consider a ramp merging scenario, where we assume that driver behavior is influenced by the drivers' internal states.
With $\theta_{\mathrm{IDM}}$, $w_l$ and $w_m$, a driver's acceleration can be determined by computing $f_{\mathrm{NIDM}}= f_l w_l + f_m w_m$, where $f_l$ and $f_m$ are calculated using \eqref{equ:idm_action} as if the driver is attending to its front neighbor and the vehicle on the on-ramp, respectively. 
As shown in \cref{fig:model_architecture}, NIDM first estimates $\theta_{\mathrm{IDM}}$, which is then used by the policy to output the driver's actions over a sequence of $T$ future time steps. 
This is in contrast to existing imitation learning settings where the policy is trained in raw action spaces and directly maps observations to actions \cite{casas2020implicit, casas2019spatially, casas2020importance, bhattacharyya2019simulating, henaff2019model, suo2021trafficsim, morton2017simultaneous}.
 
In some traffic states and for some values of internal states, the magnitude of the output from \eqref{equ:idm_action} can become exponentially large, leading to unstable training dynamics. We aid model training by constraining the space of parameters to plausible values. 
This yields more stable gradient descent dynamics at training time and ensures that an estimated internal state is consistent with the driver attributes it represents (e.g., desired speed $v_{\mathrm{des}}$ and distance $d_{\mathrm{des}}$). 
To this end, we apply clipping $\mathrm{min}\{f_{\mathrm{IDM}}, -6\}$ to the output of \eqref{equ:idm_action} and apply the ReLU operator to \eqref{equ:des_d},
\begin{equation}  
\begin{aligned} 
d_{\mathrm{des}}(v^t, \Delta v^t) = d_{\mathrm{min}} + \mathrm{ReLU}\left(T_{\mathrm{des}}v^t + \frac{v^t\Delta v^t}{2\sqrt{a_{\mathrm{max}}b_{\mathrm{max}}}}\right)
\end{aligned}
\end{equation}
We additionally define the parameters in $\theta_{\mathrm{IDM}}$ based on the generalized logistic function \cite{richards1959flexible}. For a given driver parameter $\hat \lambda$, we define the logistic function as
\begin{equation}  \label{equ:logistic_function}
\begin{aligned} 
\hat \lambda = \lambda_\mathrm{tim} + \frac{\lambda_\mathrm{agg}-\lambda_\mathrm{tim}}{1+\exp^{-\delta(\hat x)}}
\end{aligned}
\end{equation}
where $\hat x$ is specified by the neural network, $\delta$ is a tunable hyperparameter that controls the slope of the logistic function, and $\lambda_\mathrm{agg}$ and $\lambda_\mathrm{tim}$ are the parameter bounds corresponding to the most aggressive and timid drivers, respectively.
Finally, we apply a softmax activation to the attention parameters $w_l$ and $w_m$, where the operator acts as a gating function, indicating to whom a driver is attending at any given moment.
We note that all the operations in \eqref{equ:idm_action} preserve differentiability, which is required for gradient-based learning methods such as ours.
  
\subsection{NIDM Architecture}
Let $X = (a^{0:t}, x^{0:t})$ denote the vehicles' motion history, where $a^{0:t}$ and $x^{0:t}$ are the past sequence of the vehicles' joint actions and states, respectively. The future motion is denoted as $Y = (a^{t:t+T}, x^{t:t+T})$.
We extract information from $X$ and $Y$ using long short-term memory (LSTM) networks ~\cite{hochreiter1997long}. We use two LSTM-based encoders to turn $X$ and $Y$ into representations $h_x$ and $h_y$, respectively. 
To obtain a generative model of driver behavior, we employ the framework of the conditional variational auto-encoder \cite{sohn2015learning}, which encodes probability distributions in terms of a latent variable $Z$ such that 
$P(Y|X) = \int\displaylimits_{Z} P(Y|X, Z)P(Z|X)$, where $P(Y|X)$ represents the conditional distribution over future trajectories and the variable $Z$ captures the unobserved and uncertain latent factors such as a driver's disposition and attentiveness. 
Since integration over $Z$ is intractable, we exploit amortized variational inference \cite{higgins2016beta, sohn2015learning}, which introduces a neural network approximation of the posterior $Q_{\phi}(Z|h_x, h_y)$ and prior $P_{\theta}(Z|h_x)$ to reformulate the model learning problem as maximizing the evidence lower bound (ELBO) on $\mathrm{log}P(Y|X)$. 

\subsection{Training NIDM}
We train NIDM end-to-end in an imitation learning setup. 
It is important to note that since we do not consider the drivers' internal states to be known a priori, it is not possible to parameterize the probability distribution over them using approaches (e.g., Gaussian mixture models or mixture density networks \cite{schulz2019learning, morton2017simultaneous}) that rely on the availability of ground truth values for model learning.
 
For a given training example, a single sample $\hat Z \sim Q_{\phi}(Z|h_x, h_y)$ is drawn and used to infer the driver's internal state. We then roll out the policy (i.e., in closed-loop) during training to encourage the learning of a decoder $\theta_{\mathrm{IDM}} = \Phi_{\mathrm{IDM}}(Z)$ that yields improved long-term predictions. 
Starting from the vehicle's initial state, we roll out the policy and use the following discrete-time equations to propagate the vehicle's state forward in time:
\begin{align} \label{equ:agent_dynamics_1}
&
\mathrm{x}^{t+1}\approx \mathrm{x}^{t}+ v^{t}\Delta t+ \frac{1}{2} \hat a^{t}\Delta t^{2} \\
& \label{equ:agent_dynamics_2}
v^{t+1}\approx v^{t}+ \hat a^{t}\Delta t
\end{align}
where $\hat a^t$ is the driver's estimated longitudinal acceleration, $\mathrm{x}^t$ is the vehicle's longitudinal position, $v^t$ is the vehicle's speed, and $\Delta t$ is a small time interval. Note that since the transition model \eqref{equ:agent_dynamics_1} is composed of differentiable elementary operations,
gradients remain well-defined---gradient information can be back-propagated through the simulated trajectory and the driver's internal state. 

In this paper, we have followed existing works \cite{monteil2015real, ward2017probabilistic, sunberg2020improving, bhattacharyya2020online} in making the simplifying assumption that $\theta_{\mathrm{IDM}}$, which represents drivers' disposition, does not vary during a policy rollout. In some cases, this assumption may not be realistic, given that human behavioral patterns are often inconsistent across time.
Relaxation of this assumption would require us to track the change in driver disposition over time, which has been left to future work. 
However, the same assumption does not apply to driver attention, as we consider a driver's attention to be dependent on the traffic state (e.g., the position and acceleration profile of a vehicle on the on-ramp). As illustrated in \cref{fig:model_architecture}, we estimate driver attention sequentially at every step of the policy rollout.

As the training procedure, we adapt the variational learning objective of CVAE \cite{sohn2015learning} to maximize the ELBO of $\mathrm{log}P(Y|X)$. 
In total, the loss consists of three terms $\mathcal{L}_{\mathrm{Total}} = \mathcal{L}_{a} + \mathcal{L}_\mathrm{x} + \beta\mathcal{L}_{\mathrm{KL}}$, where the first two terms are the
vehicle acceleration and position error, respectively, between the ground truth values and those generated by the model. The last term is the Kullback-Liebler (KL) divergence weighted by the scalar parameter $\beta$.
Concretely, the loss terms are defined as: 
\begin{align} \label{equ:loss_func}
&
\mathcal{L}_{a} =\frac{1}{T} \sum\limits_{i=0}^{T}L_{\delta}(a^{t+i}-\hat a^{t+i}) \\
&
\mathcal{L}_{\mathrm{x}} =\frac{1}{T} \sum\limits_{i=0}^{T}L_{\delta}(\mathrm{x}^{t+i+1}-\mathrm{\hat x}^{t+i+1}) \\
&
\mathcal{L}_{\mathrm{KL}} = \textrm{KL}(Q_{\phi}(Z|h_x, h_y)\;||\;P_{\theta}(Z|h_x))
\end{align}
where $L_{\delta}$ stands for the Huber loss. 

\begin{figure}
	\centering
	\includegraphics[trim={0.3cm 0.2cm 0.2cm 0.2cm}, clip=true,
    width=\linewidth]{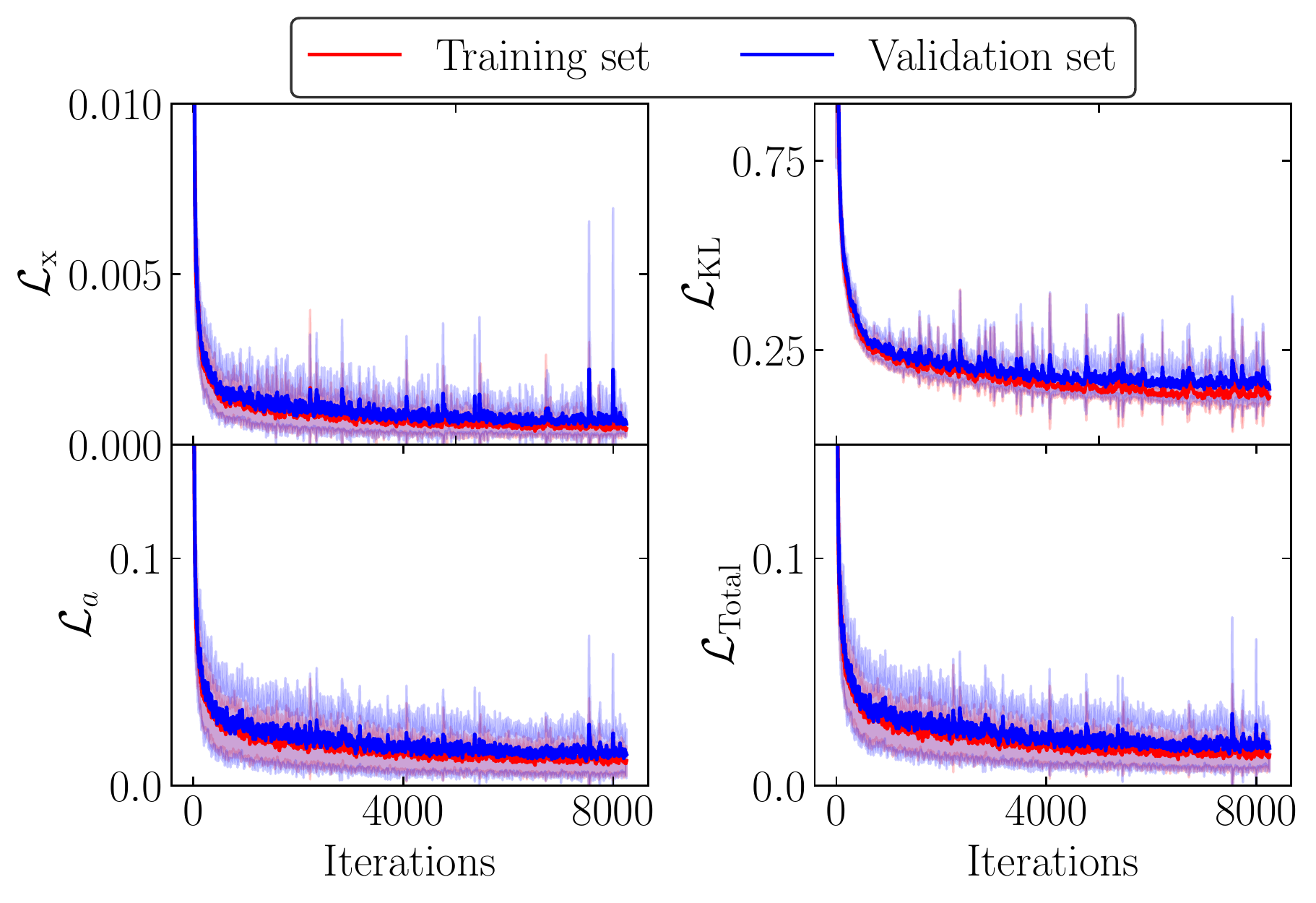}
	\caption{Convergence of our model training on the training (red) and validation (blue) data. Loss statistics are obtained from 10 differently seeded NIDM models and the curves are smoothed over 10 iterations.}
	\label{fig:loss_plot}
\end{figure}

We implement NIDM using the TensorFlow library \cite{abadi2016tensorflow} and train it with the Adam optimizer \cite{kingma2014adam} with a learning rate of $0.001$. In addition to hyperparameter annealing, we found standardization of both input and target values leads to faster convergence. We ran experiments with 3, 6, and 9 latent variables, and found that for our dataset, all model variants provide comparable validation errors upon convergence. We use 6 latent variables for the rest of our analyses. 
We found that setting $\delta = 4/(\lambda_\mathrm{agg}-\lambda_\mathrm{tim})$ in \eqref{equ:logistic_function} provides a good solution quality. 
We also found that setting $\beta=0.02$ yields a good trade-off between the loss terms, without one loss component overpowering the others.
The loss statistics are shown in \Cref{fig:loss_plot}. As shown, all the loss terms decrease rapidly before converging to a relatively low value for both the training and validation sets. 
 
\section{Simulation Setup}\label{sec:Simulation_Setup}
\subsection{Driving Strategy}

Within the simulation environment, the longitudinal acceleration of vehicles is determined by IDM and their merge decisions are determined by MOBIL \cite{kesting2007general}, which is a rule-based strategy that determines a vehicle's decision with the goal of maximizing the longitudinal acceleration for the vehicle and its neighboring cars. A decision to merge is considered if the induced acceleration of the immediate rear car on the main lane fulfills the safety criterion $a_n > b_{\mathrm{safe}}$. Given that the safety criterion is met, a lane merge is performed if,   
\begin{equation}  
\begin{aligned} 
\tilde a_{c} - a_{c} + C((\tilde a_n-a_n)+(\tilde a_o-a_o)) > a_{th}
\end{aligned}
\end{equation}
where $a_c$, $a_n$ and $a_o$ are the actions of the vehicle making the merge decision, the rear car in the new lane and the rear car in the current lane, respectively. The accelerations with tildes are calculated as if a lane merge has been initiated. The  parameter $C \in [0, 1]$ is a politeness factor which represents the degree to which the speed gains and losses of the surrounding vehicles are valued over that of the vehicle wanting to merge. 

\subsection{Driver Disposition}

\def\arraystretch{1.2}
\begin{table}
\begin{center}
\caption{IDM AND MOBIL DRIVER PARAMETER RANGE.}
\centering
\hrule
\hrule
\begin{tabular}{m{3.8cm} m{1cm} >{\centering\arraybackslash}m{1cm} >{\centering\arraybackslash}m{1cm}} 
\label{table:idmmovil_params}
Driver Parameter &  & Aggressive ($\lambda_\mathrm{agg}$) & Timid ($\lambda_\mathrm{tim}$)\\ 
\hline
Desired speed (\SI{}{\meter\per\second}) & $v_{\mathrm{des}}$ & 25 & 15 \\ 
Desired time gap (\SI{}{\second}) & $T_{\mathrm{des}}$ & 0.5 & 2 \\
Min separation distance (\SI{}{\meter}) & $d_{\mathrm{min}}$ & 1 & 5 \\
Max acceleration (\SI{}{\meter\per\second\squared}) & $a_{\mathrm{max}}$ & 4 & 2 \\
Max deceleration (\SI{}{\meter\per\second\squared}) & $b_{\mathrm{max}}$ & 4 & 2 \\ 
Safe braking (\SI{}{\meter\per\second\squared}) & $b_{\mathrm{safe}}$ & -5 & -3 \\
Acceleration threshold & $a_{th}$ & 0 & 0.2 \\
\hline
\end{tabular}
\end{center}
\end{table}

The ranges of possible values for the driver parameters are listed in \cref{table:idmmovil_params}. At the start of a driving episode, we populate the road with drivers that have aggressiveness levels ranging from the most aggressive to the most timid. The roadway consists of a \SI{500}{\meter} main road and a \SI{100}{\meter} on-ramp. To expose the model to a diverse range of traffic conditions, we populate the road with $n= \{n_1, n_2, ..., n_N\}$ vehicles where $N\in_R \{4, 5, 6, 7\}$ is the total number of vehicles on the road for a given episode.

The driver parameters are assigned as follows. 
For each driver, we first draw a sample from a uniform distribution $\psi\sim U(0, 1)$,
where $\psi$ represents the driver's \textit{aggressiveness level}. Drivers with larger $\psi$ express more aggressive tendencies, such as driving close to their front neighbor (tailgating) or not being cognizant of the vehicles on the on-ramp. Given $\psi$, each driver parameter in $\theta_{\mathrm{IDM}}$ is considered to be distributed according to a Beta distribution $\mathrm{Beta}(\alpha, \beta)$ with the shape factors~$\alpha=\phi\psi$ and $\beta = \phi(1-\psi)$, where $\phi$ is a precision coefficient that alters the variance of a given beta density. 
For each driver parameter $\lambda$, we draw a sample $\tilde \psi\sim \mathrm{Beta}(\alpha, \beta)$ to calculate the parameter value $\lambda = \lambda_\mathrm{tim} + \tilde \psi(\lambda_\mathrm{agg}-\lambda_\mathrm{tim})$. This way, drivers with a wide range of parameter values can be created while the parameters for an individual driver remain correlated, hence maintaining the notion of driver aggressiveness.
In our experiments, we use a precision value $\phi=15$. 
 
\subsection{Driver Attention}

We adapt the Cooperative Intelligent Driver Model (C-IDM) proposed in \cite{bouton2019cooperation} for modeling driver attention. C-IDM uses estimates of time to reach the merge point $(TTM)$ for the vehicle on the main road $(TTM_a)$ and the vehicle on the on-ramp $(TTM_b)$.
Within the simulator, the attention of a vehicle is assigned as follows: 
\begin{itemize}
    \item If  $TTM_b < C \cdot TTM_a$, the vehicle on the main lane follows IDM by considering the projection of the merging vehicle on the main lane as its front neighbor, therefore exhibiting a yielding behavior. 
    \item If $TTM_b \geq C \cdot TTM_a$, the vehicle on the main lane follows the standard IDM, therefore exhibiting a passing behavior. 
    \item In the absence of any vehicles on the on-ramp, the vehicle on the main lane follows the standard IDM.
    \item In the case that a vehicle on the on-ramp decides to merge but $TTM_b \geq C \cdot TTM_a$, the vehicle on the main lane yields to the merging car when $a_n < b_{\mathrm{safe}}$ to avoid a collision.
\end{itemize}

\subsection{Training Features} \label{sec:training_features}
We train NIDM and other evaluation baselines on a synthetic dataset
collected from the simulated ramp merging scenario. 
The synthetic dataset contains 500 driving episodes which amount to roughly three hours of driving by 1500 drivers, each with unique driver parameters. 
A set of features are chosen to represent the local traffic context from each driver's viewpoint. In addition to the joint vehicle state and action, a feature vector contains the Euclidean distance of the on-ramp vehicles to the merging point
and a Boolean feature to indicate whether there is a vehicle present on the on-ramp. We set the feature values of the missing vehicles to dummy values equal to their mean value across all the training examples.  

\section{EXPERIMENTS AND RESULTS}

\subsection{Baseline Policies}

We compare the performance of NIDM to the following~baselines:
\begin{itemize}
        \item Multilayer perceptron (\textbf{MLP}) policy. The MLP is a 4-layer, fully connected architecture composed of ReLU activation functions similar to that proposed in \cite{schulz2019learning}. 
        \item \textbf{LSTM} policy. This is similar to MLP but it also maintains an internal state that conditions the policy on the vehicles' motion history.
        \item \textbf{Latent-MLP} policy. This is a policy from \cite{morton2017simultaneous}, which was originally evaluated on a straight roadway. 
        \item \textbf{CVAE}. This is a simplified setting of our model, where we remove the IDM layer so that the decoder directly maps its inputs to driver actions. 
\end{itemize}
The MLP receives the current traffic state as input to parameterize the distribution over driver actions, with the training objective of maximizing the log-likelihood of true driver actions.
Similar to our approach, Latent-MLP can learn a latent, low-dimensional representation of driving trajectories. Latent-MLP uses a normal prior distribution $\mathcal{N}(0, 1)$ (while the prior in NIDM is approximated by a neural network) and GMMs are used to characterize the distribution over driver actions. 
For both NIDM and CVAE models, we use the same experimental protocol, where the dataset is split into 70\% training and 30\% validation. With a time interval of $\Delta t = \SI{0.1}{\second}$, we extract 8 seconds (80 time steps) of trajectories, using the first 3 seconds as motion history to predict the next 5 seconds. The same dataset is used for other baselines, although their training involves predicting actions for one step at a time as opposed to generating full trajectories through policy rollouts.

\subsection{Evaluation Procedure}
The models are evaluated within the same simulation environment we use to collect the training data. Each simulation episode lasts for \SI{10}{\second}, corresponding to 100 time steps at a simulation frequency of \SI{10}{\Hz}. 
The procedure we use to generate vehicle trajectories proceeds as follows:
\begin{enumerate}
    \item For a given episode, the road is populated with simulated drivers with randomly assigned initial speeds, initial positions and driver parameters.
    \item The simulation is run for \SI{3}{\second} to obtain sufficient motion history for feeding each driver policy.
    \item After \SI{3}{\second}, the policies take over to predict the longitudinal acceleration of each vehicle on the main road.
    \item Equation \eqref{equ:agent_dynamics_1} is applied to estimate the vehicles' position at the next time step. 
    \item In the new traffic state, a feature vector (see \cref{sec:training_features} for details) for every driver is obtained to sequentially feed their policies.
    \item Steps 3, 4 and 5 are repeated to propagate the traffic state forward in time.
\end{enumerate}
 
Upon the completion of the procedure above, we can compare the generated vehicle trajectories to the ground truth trajectories for the performance evaluation and comparison of different policies. We note that although the vehicles share the neural network parameters of a single policy, the policies are able to produce a diverse range of driving behaviors as they receive different observation sequences.   

\subsection{Evaluation Metric}

We use Root-Weighted Square Error (RWSE) as the performance metric, which captures the deviation of the model’s probability mass from ground truth  trajectories~\cite{morton2017simultaneous}. We note that data likelihood, which is a measure for the goodness-of-fit of the model to data, is not a sufficient metric to assess model performance, as it only captures similarity at the level of individual data points.
The RWSE for $m$ trajectories, $n$ sampled traces per true trajectory, and for the predicted variable $r$ is:
\begin{equation}
\begin{aligned} 
\mathrm{RWSE}^t = \sqrt{\frac{1}{mn}\sum_{i=1}^{m}\sum_{j=1}^{n}(r^{t}_{(i)}-\hat{r}^{t}_{(i,j)})^2\ } 
\end{aligned}
\end{equation}
where $\hat{r}^{t}_{(i,j)}$ is the predicted variable at time $t$ and under the sample $j$. For the quantitative results presented in this section, we generate $m=210$ trajectories and $n=10$ sampled traces, resulting in a total of 2100 policy rollouts. 
\subsection{Qualitative Evaluation}
 
\begin{figure} 
	\centering
	\subfloat[CVAE\label{fig:neural_latent}]{
    	\includegraphics[width=0.43\linewidth, trim={0.3cm 0.2cm 0cm 0.7cm}, clip=true]{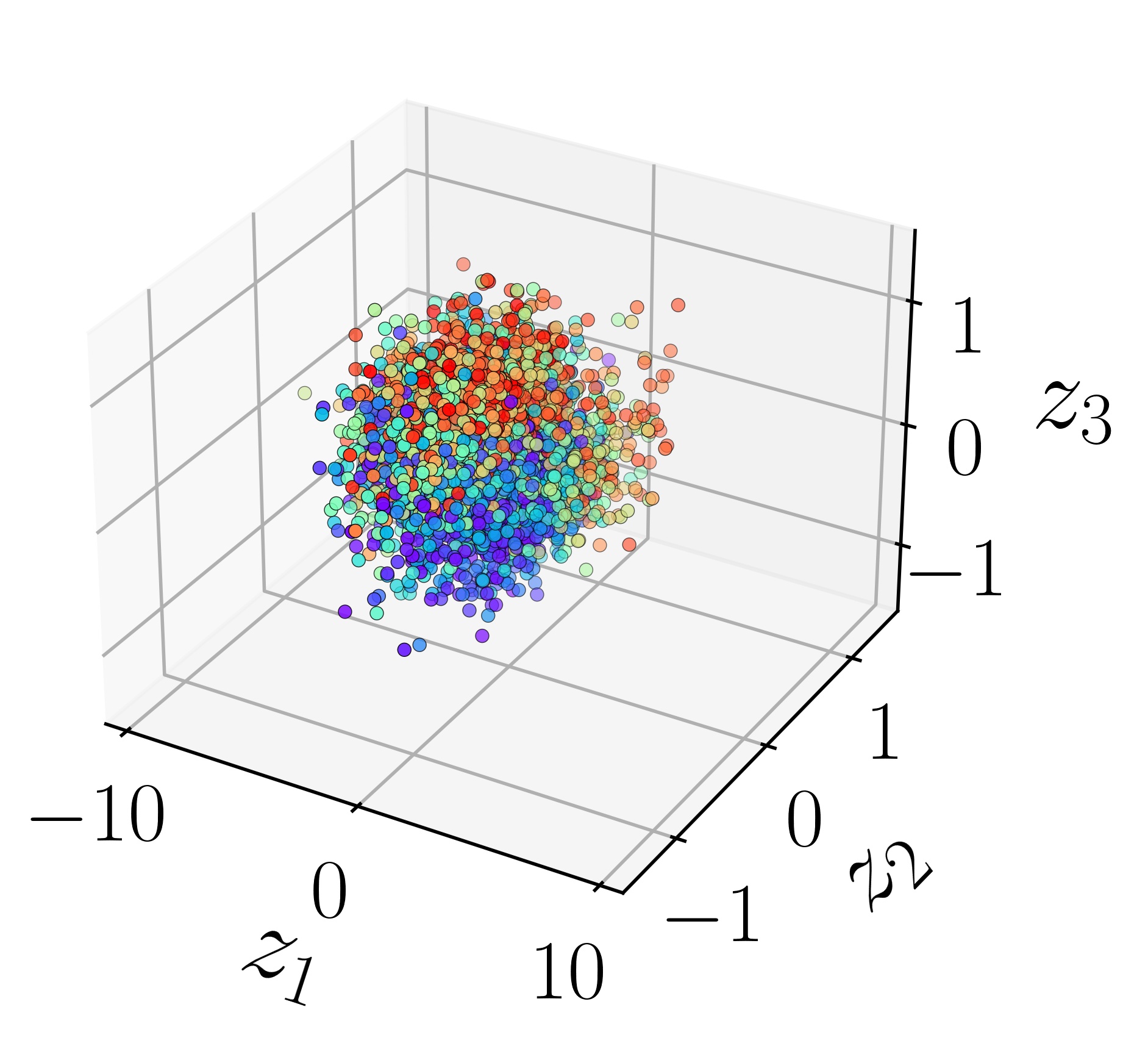}}
	\subfloat[NIDM\label{fig:NIDM_latent}]{
    	\includegraphics[width=0.53\linewidth, trim={0.6cm 0.2cm 0cm 0.7cm}, clip=true]{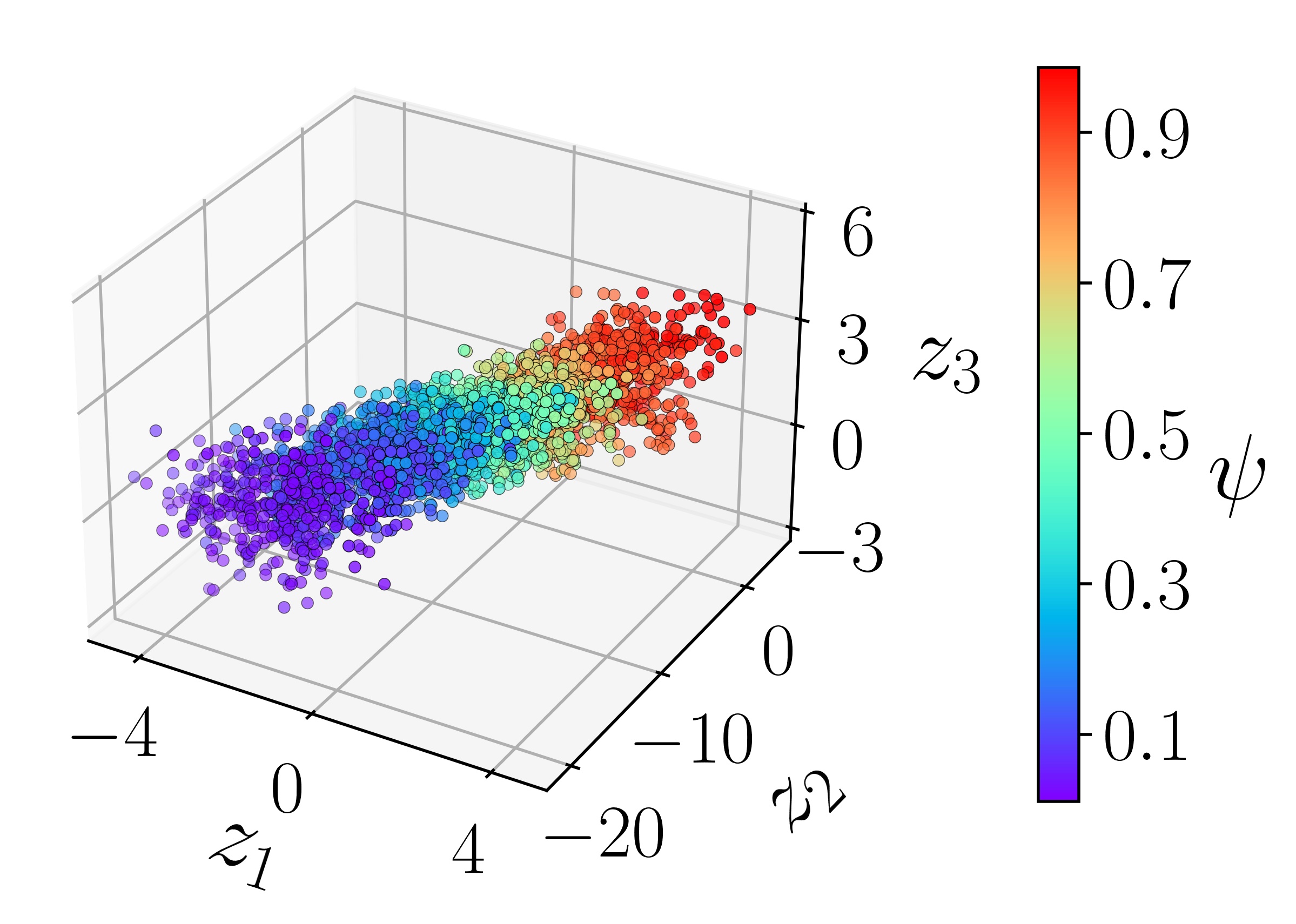}}
	\caption{Visualization of the three-dimensional latent space with coloring according to driver aggressiveness.}
	\label{fig:latent}
\end{figure}

\begin{figure*} 
	\centering
	\subfloat[\label{fig:example_prediction_a}]{
    	\includegraphics[width=0.325\linewidth, trim={0cm 0.4cm 0cm 0.cm}, clip=true]{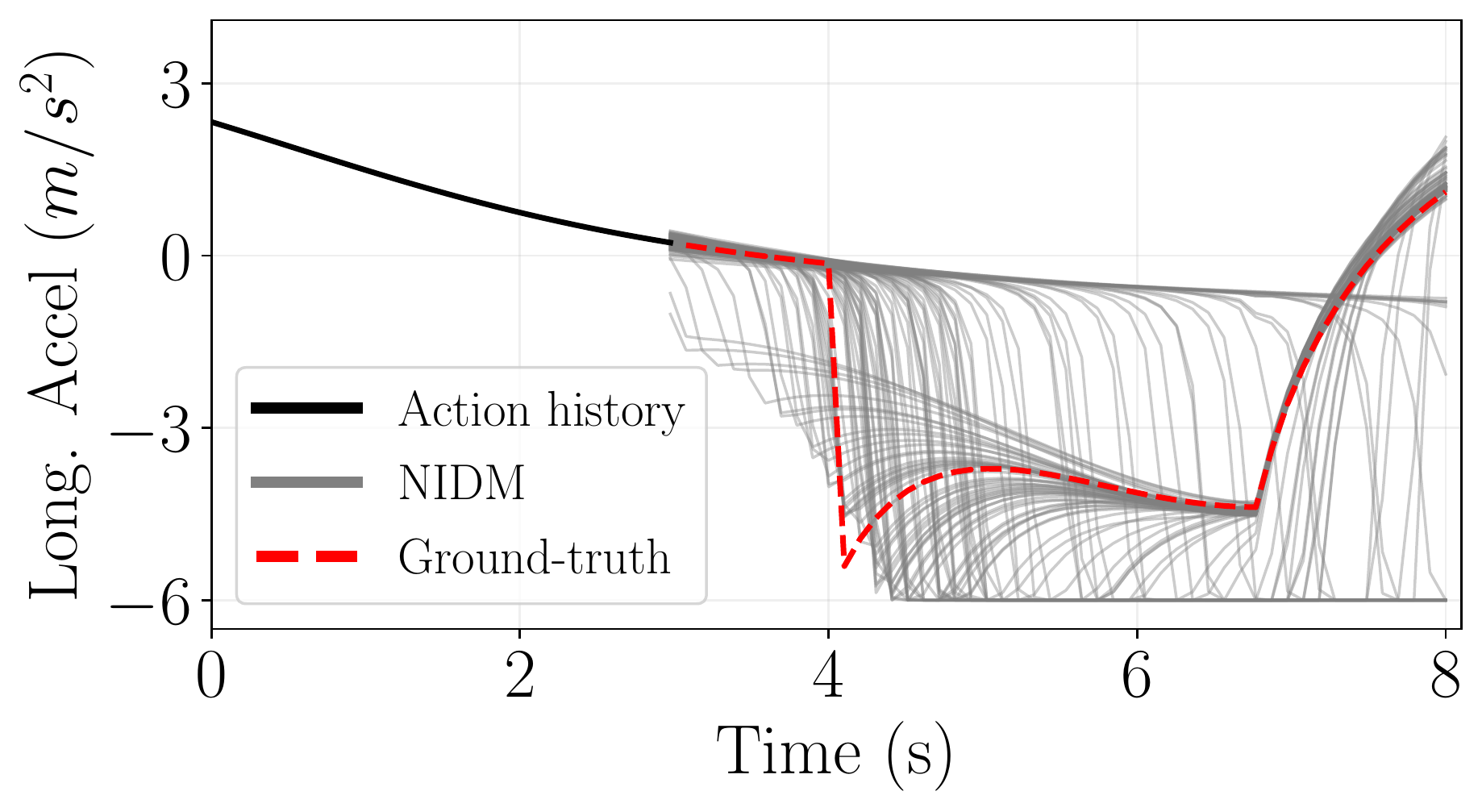}}  
	\subfloat[\label{fig:example_prediction_b}]{
    	\includegraphics[width=0.325\linewidth, trim={0cm 0.4cm 0cm 0.cm}, clip=true]{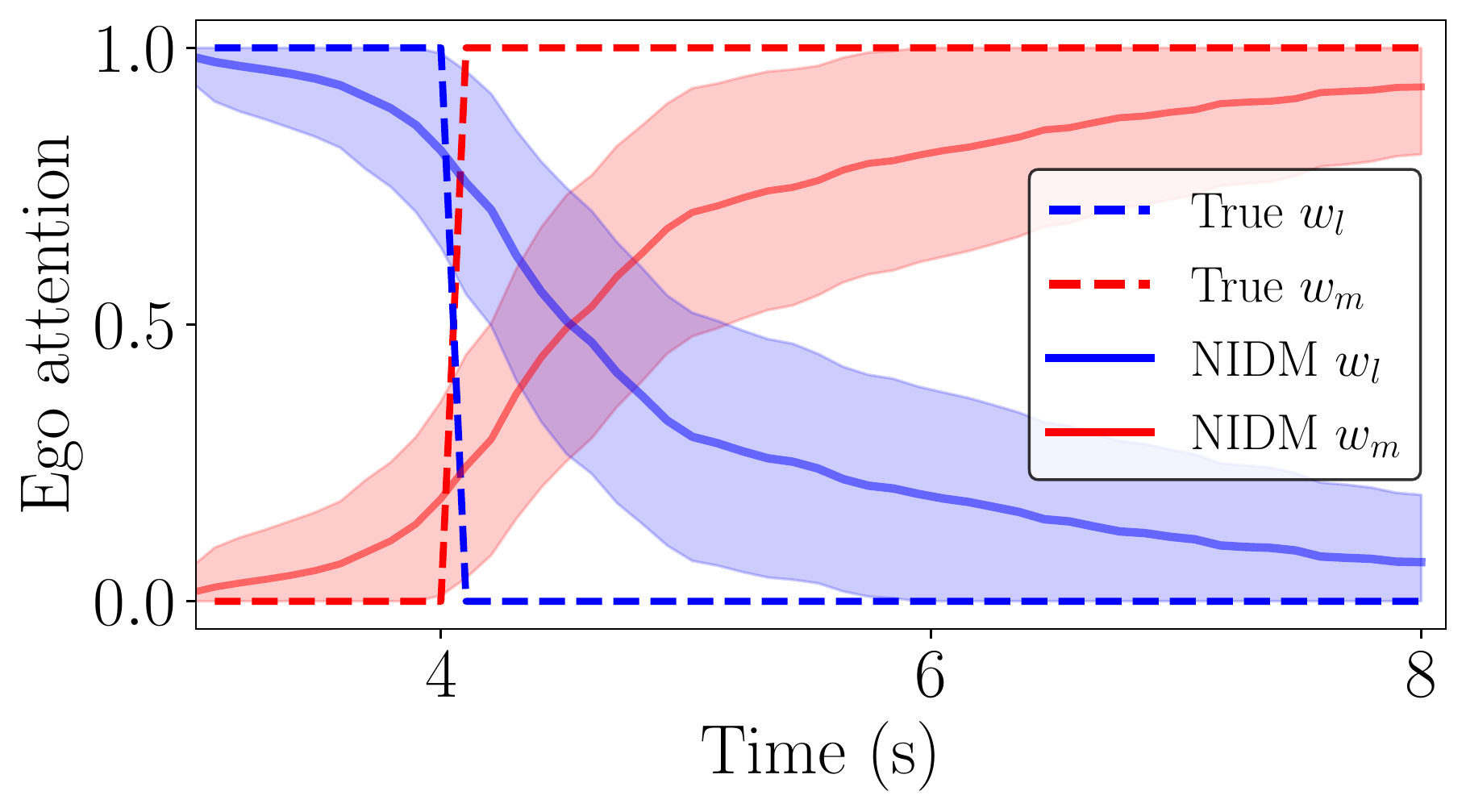}}  	
	\subfloat[\label{fig:example_prediction_c}]{
    	\includegraphics[width=0.325\linewidth, trim={0cm 0.4cm 0cm 0.cm}, clip=true]{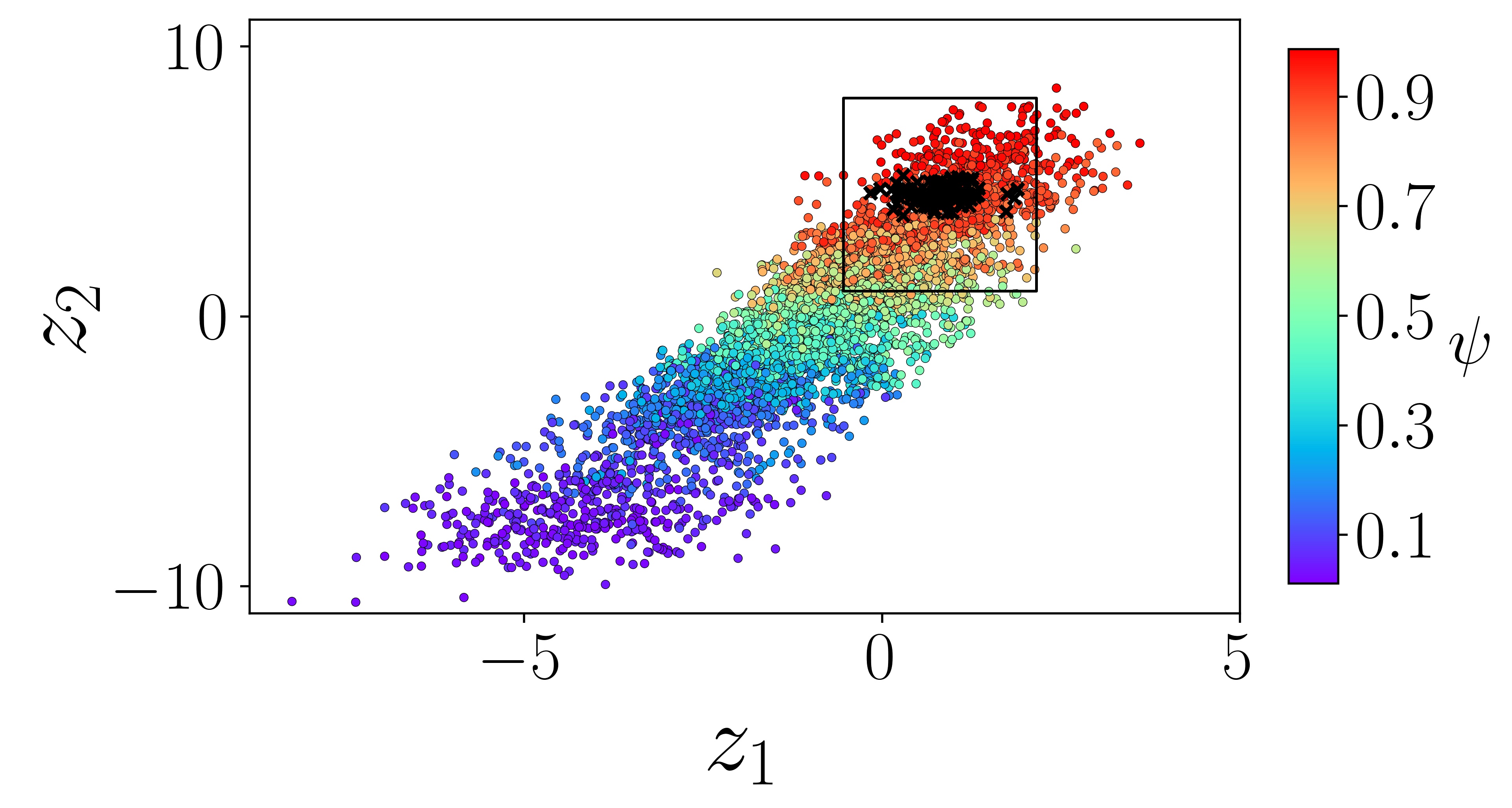}} \\[-0.5ex]
	\subfloat[\label{fig:example_prediction_d}]{
	\includegraphics[width=0.99\linewidth, trim={0cm 0.3cm 0cm 0.cm}, clip=true]{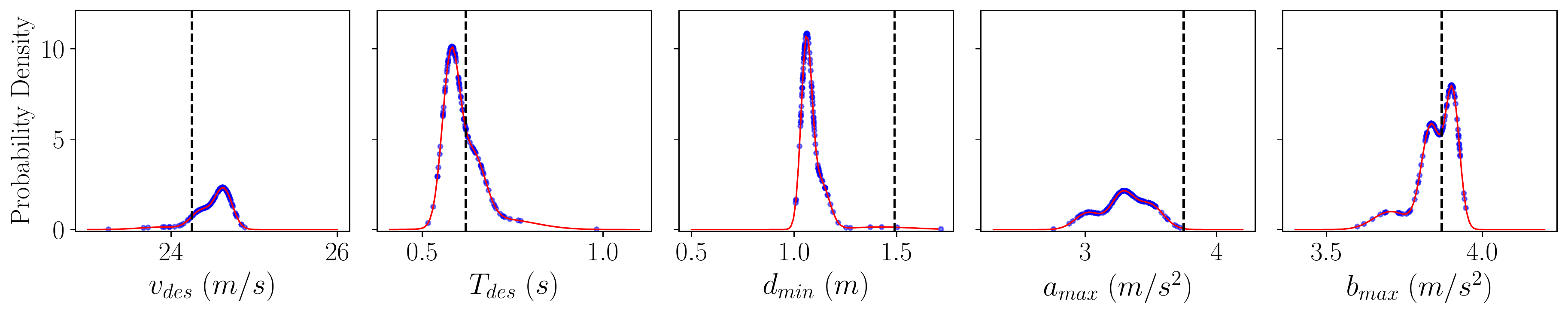}}
	\caption{Sample predictions (100 samples) for a scenario in which an on-ramp vehicle decides to merge in front of a vehicle on the main road. (a) True and predicted driver action profiles. (b) Anticipated driver attention with its mean and standard errors. (c) Visualization of the two-dimensional latent space for the validation set with coloring according to the ground truth driver aggressiveness; the black crosses are samples from the latent space that were generated for this specific scenario. 
	(d) Sampled values for the driver parameters~$\theta_{\mathrm{IDM}}$ (marked blue) fitted with Gaussian mixtures (marked red) for clarity.}
	\label{fig:example_prediction}
	
\end{figure*}
 
\subsubsection{Latent Space Visualization} 

\cref{fig:latent} shows the projection of three of the six latent variables for NIDM and CVAE.
The plot was generated using 5000 trajectories that were randomly extracted from the validation dataset and the points in the plot are colored according to driver aggressiveness (ranging
from purple (most timid) to red (most aggressive)). 
It should be noted that the latent samples shown are from the prior network $P_{\theta}(Z|h_x)$. We observe that NIDM successfully clusters together drivers according to their aggressiveness. 
Compared to CVAE, NIDM appears to significantly improve the information content of the latent embedding, as illustrated by a more semantic latent representation. Intuitively, the lowest training loss can be achieved by arranging the nearby trajectories in data space (which carry cues about the drivers' internal state) closer together in the latent space. 
We conjecture that the discrepancy between the two models is caused by the decoder’s ability in the CVAE model to sequentially map its inputs directly to driver actions over time. In contrast, for NIDM, the parameter set $\theta_{\mathrm{IDM}}$ is inferred once at the start of a given policy rollout and used repeatedly to propagate the vehicle's state forward in time. This constraint on NIDM establishes a strong driving factor for shaping the latent space,
encouraging the learning of an informative embedding by exploiting the gradient paths through the state transitions over time.
 
\subsubsection{Ramp Merging Scenario} 

\cref{fig:example_prediction} gives a qualitative impression of the NIDM's predictions for an example scenario.
The scenario involves a vehicle merging in front of a relatively aggressive driver with an aggressiveness level of $\psi=0.9$.
Given \SI{3}{\second} of motion history as input, NIDM is tasked with estimating the unknown parameters $\theta_{\mathrm{IDM}}$ and predicting the driver's future behavior. It is important to note that the input to NIDM contains no explicit information (such as labels) about the driver's internal state.  
After about \SI{4}{\second}, the driver starts to attend to the on-ramp vehicle, as indicated by the sharp deceleration actions shown in \cref{fig:example_prediction_a}. This behavior is typical of aggressive drivers in our traffic simulator, who would not brake until a car merges in front of them, by which time the available gap has closed so much that abrupt braking becomes necessary to avoid a collision.
We observe that NIDM has correctly anticipated the driver's hard braking (marked gray), with the model predictions being distributed around the true driver actions (marked red). 

\cref{fig:example_prediction_d} shows the estimated driver parameter values, where the true values are marked by the dashed vertical lines.
The generated samples are colored in blue, which we fit with three-mode Gaussian mixtures using the expectation-maximization (EM) algorithm to aid with visualization. 
We observe that NIDM is able to correctly identify the unknown parameters $v_{\mathrm{des}}$, $T_{\mathrm{des}}$ and $b_{\mathrm{max}}$, as indicated by the large probability mass over the true parameter values. There are larger probability mass offsets for parameters $d_{\mathrm{min}}$ and $a_{\mathrm{max}}$.
We have found that these offsets can be justified by the very nature of the IDM's structure in \eqref{equ:idm_action}, which depending on the traffic state, can be more or less sensitive to different driver parameters. The output of \eqref{equ:idm_action} is most sensitive to $v_{\mathrm{des}}$ and $T_{\mathrm{des}}$, hence the model is severely penalized for having poor estimates for these parameters. Conversely, $d_{\mathrm{min}}$ is only relevant in a traffic jam situation, and hence has little impact on the output of \eqref{equ:idm_action} in this traffic instance. As such, the model does not incur large loss penalties for making poor estimates of $d_{\mathrm{min}}$. 
Intuitively, this result is to be expected, since we do not explicitly train the model to predict the driver parameter values (as they are considered to be unknown). Instead, the model's objective while training is to minimize the loss function $\mathcal{L}_{\mathrm{Total}}$, towards which inferring the unknown $\theta_{\mathrm{IDM}}$ is merely an intermediate step. 

In \cref{fig:example_prediction_c}, a visualization of the latent space for NIDM is shown, where the black crosses are samples for this specific scenario. 
By inspecting the latent space, we can observe that the driver expresses aggressive driving tendencies. With the addition of the estimated driver attention values shown in \cref{fig:example_prediction_b}, we can interpret how the anticipated driver accelerations were derived. 

\subsection{Quantitative Evaluation}

\begin{figure} 
	\centering
	\includegraphics[width=\linewidth]{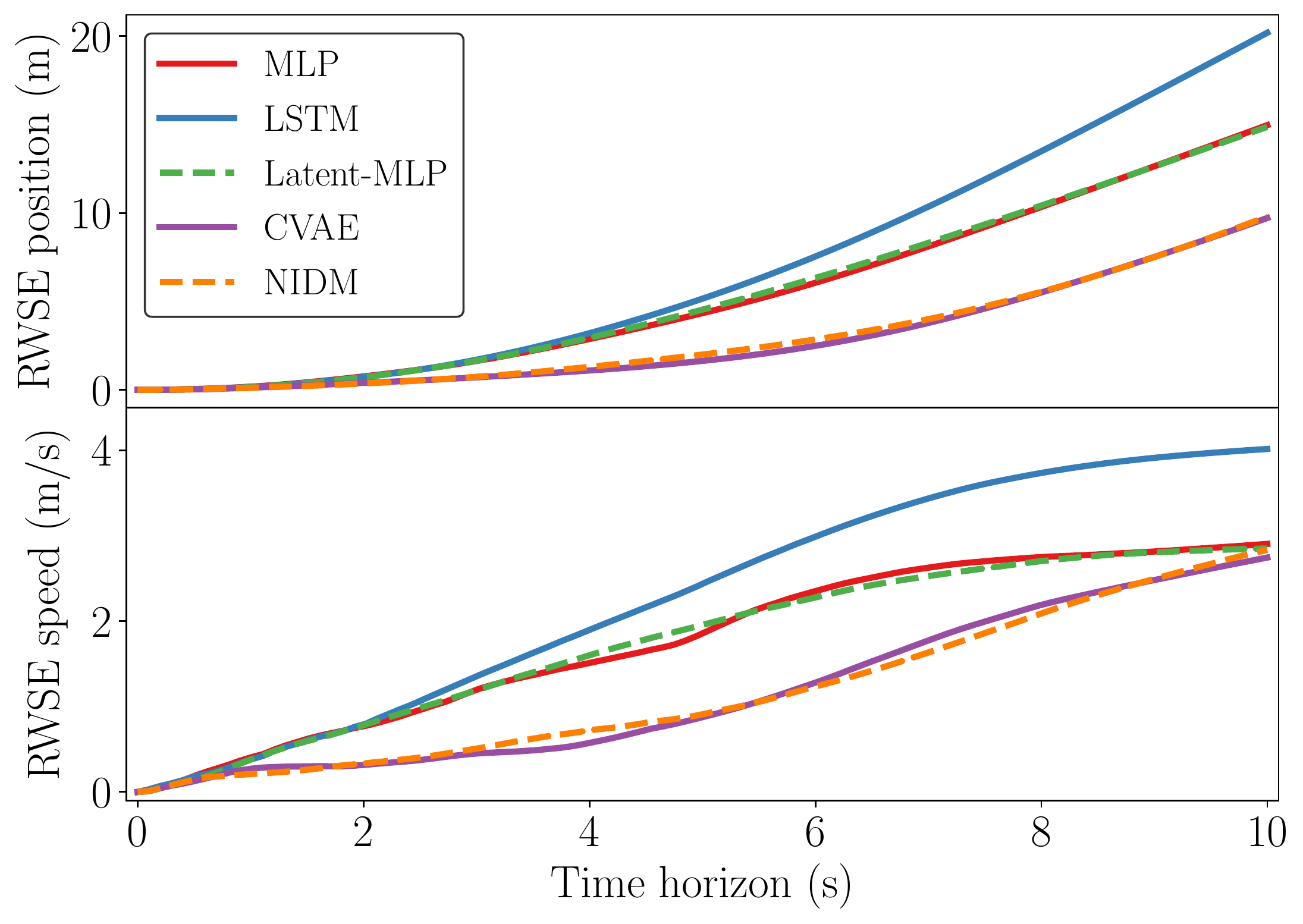}
	\caption{Root weighted square error values for the trajectories generated through 2100 policy~rollouts.}
	\label{fig:rwse}
\end{figure}
 
\begin{figure} 
	\centering
	\includegraphics[width=0.98\linewidth]{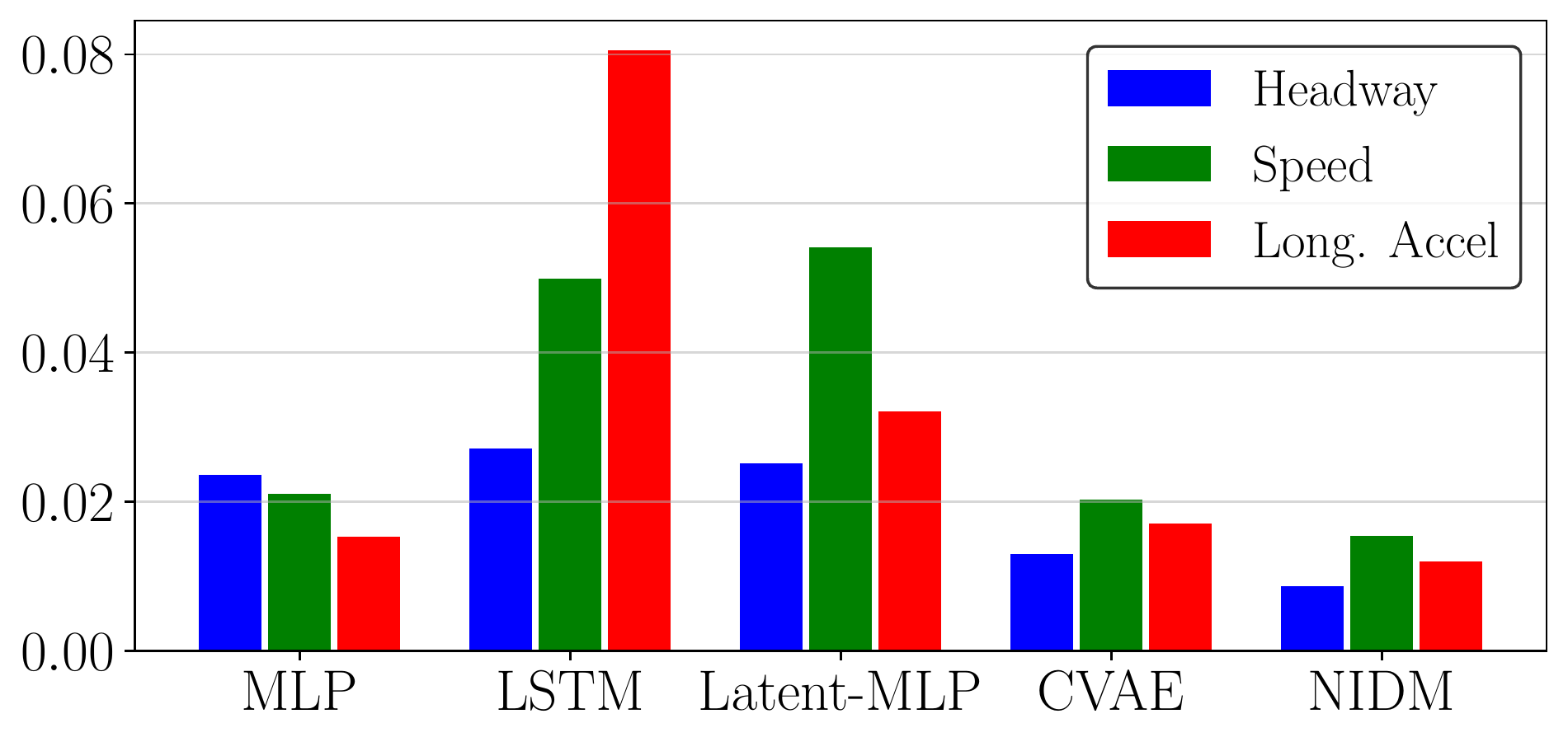}
	\caption{Summary of KL divergence values for trajectories generated using each model.}
	\label{fig:kl_bar_chart}
\end{figure}

\cref{fig:rwse} shows the RWSE values against the prediction horizon for vehicle speed and position. NIDM and CVAE yield a greater generalization capability, as indicated by the lower RWSE values. The performance discrepancy becomes more pronounced further into the future, where both NIDM and CVAE are found to be less prone to degrading accuracy compared to other baselines.
In \cref{fig:kl_bar_chart} we use KL-divergence to quantify the degree of similarity between the distributions over the emergent vehicle state and action values encountered during the policy rollouts.
NIDM yields lower KL-divergence values, with CVAE showing comparable results. 

In \cref{table:collision_eval}, we report the resulting collision counts and collision rates from the policy rollouts. 
We find that NIDM has the lowest collision rate out of the evaluated baselines, followed by CVAE with a collision rate of 2.4\%. 
This result empirically demonstrates
that NIDM is less likely to lead vehicles into collisions.
A plausible explanation for NIDM's lower collision rate is that the IDM's dynamic equation has been designed to express stability properties that constrain the long-term behavior of the NIDM drivers. As such, prediction errors are prevented from severely cascading into the future due to distribution shift.
 
\begin{table}
\begin{center}
\caption{EVALUATION RESULTS FOR RAMP MERGING SCENARIO}
\hrule
\hrule
\centering
\begin{tabular}{m{2.2cm}
>{\centering\arraybackslash}m{0.5cm} >{\centering\arraybackslash}m{0.5cm} >{\centering\arraybackslash}m{1.cm}>{\centering\arraybackslash}m{0.5cm}>{\centering\arraybackslash}m{1.cm}} \label{table:collision_eval}
& MLP & LSTM & Latent-MLP & CVAE & NIDM\\ [0.5ex] 
\hline
Collision Count & 132 & 193 & 220 & 50 & \textbf{19} \\ 
Collision Rate (\%) & 6.3 & 9.1 & 10.5 & 2.4 & \textbf{0.9} \\
\hline
\end{tabular}
\end{center}
\end{table}

\section{conclusions}

We have introduced a new approach for learning a driver model with strong inductive biases which are imposed through an expert-informed dynamic structure and that add a degree of interpretability to the black-box nature of deep networks. We show that the model can accurately replicate ground truth driver behavior over longer horizons than those experienced by the model during training. Additionally, our model includes a generative component to capture uncertainty and generate a diverse range of driver behaviors. 

The use of IDM in this work was a design choice; in practice, other forms of inductive biases can be utilized in its place, as long as they are amenable to gradient-based optimization. 
While our results demonstrate the potential of the proposed approach, it remains to be explored to what degree the performance gains extend to real-world driving datasets. 
We aim to use the proposed model for model-based planning, framing autonomous driving as a Partially Observable Markov Decision Process (POMDP) \cite{sunberg2020improving} with drivers' internal states modeled as the partially observable state variables.
The source code for the simulator and the results associated with this paper are available at https://github.com/saArbabi/DriverActionEstimators.

\section*{Acknowledgment}
This work was supported by Jaguar Land Rover and EPSRC projects ROSSINI (EP/S016317/1) and TASCC (EP/N01300X/1).
 
\bibliography{main}
\bibliographystyle{IEEEtran}
\end{document}